83

# Image Segmentation by Using Thershod Techniques

Salem Saleh Al-amri[1], N.V. Kalyankar[2] and Khamitkar S.D [3]

**Abstract**—*This paper attempts to undertake the study of segmentation image techniques by using five threshold methods as Mean method, P-tile method, Histogram Dependent Technique (HDT), Edge Maximization Technique (EMT) and visual Technique and they are compared with one another so as to choose the best technique for threshold segmentation techniques image. These techniques applied on three satellite images to choose base guesses for threshold segmentation image.*

—————————— ◆ ——————————

## 1 INTRODUCTION

Segmentation algorithms are based on one of two basic properties of intensity values discontinuity and similarity. First category is to partition an image based on abrupt changes in intensity, such as edges in an image. Second category is based on partitioning an image into regions that are similar according to predefined criteria. Histogram Threshold approach falls under this category. This paper taken the study the second category (threshold techniques) in this case their are more studies take this subject can be give some of these studies briefly. Threshold segmentation techniques can be grouped in three different classes:

1st.Local techniques are based on the local properties of the pixels and their neighborhoods .2nd Global techniques segment an image on the basis of information obtain globally (e.g. by using image histogram; global texture properties ). 3rd Split, merge and growing techniques use both the notions of homogeneity and geometrical proximity in order to obtain good segmentation results. Finally image segmentation, a field of image analysis, is used to group pixels into regions to determine an image's composition. [1][2].

They are proposed an auto-adaptive threshold method of two-dimensional (2-D) histogram based on multi-resolution analysis (MRA), decreasing the calculation complexity of 2-D histogram whereas improving the searching precision of multi-resolution threshold method. Such method originates from the extraordinary segmentation effects achieved by 2-D histogram threshold segmentation method through the spatial correlation of gray level and the flexibility as well as efficiency of the threshold searching of multi-resolution threshold segmentation method. Experiments results demonstrated that such method can obtain segmentation results similar with the exhaustive 2-D histogram method, whereas the calculation complexity decreases exponentially with the increase of resolution level [3].

The image threshold problem is treated as an important issue in image processing, and it can not only reduce the image data, but also lay a good foundation for succulent target recognition and image understanding. Character of global threshold segmentation and local threshold was analyzed in image segmentation. A new threshold statistic iterative arithmetic is presented to overcome the direct worth method in threshold, aiming at some lighting asymmetry and the abrupt a blemish for, or bigger arithmetic figure in ratio in a variety in gray of background image. Statistics iterative threshold segmentation, based on image gray histogram and Gauss statistics distributing, obtain the theory expression of statistics iterative method and the best worth threshold method and steps. Aviation image was threshold segmentation using statistic iterative arithmetic, histogram technique and adaptive method respectively. Compared three threshold results, it shows that statistic iterative method greatly improved the anti-noise capability of image segmentation and had the good result to the image of the worth and not easy to segment in full value threshold method [4].

Fuzzy C-means its improvement methods algorithm and strategies for remote sensing image segmentation can offer less iterations times to converge to global optimal solution. At the same time, it has good stability and robustness. Its good effect of segmentation can improve accuracy and efficiency of remote sensing image threshold segmentation [5].

Threshold techniques can be categorized into two classes: global threshold and local (adaptive) threshold. In the global threshold, a single threshold value is used in the whole image. In the local threshold, a threshold value is assigned to each pixel to determine whether it belongs to the foreground or the background pixel using local information around the pixel. Because of the advantage of simple and easy implementation, the global threshold has been a popular technique in many years [6][7][8].

## 2 THRESHOLDS

Threshold is one of the widely methods used for image segmentation. It is useful in discriminating foreground from the background. By selecting an adequate threshold value T, the gray level image can be converted to binary image. The binary image should contain all of the essential information about the position and shape of the objects of interest (foreground). The advantage of obtaining first a binary image is that it reduces the complexity of the data and simplifies the process of recognition and classification. The most common way to convert a gray-level image to a binary image is to select a single threshold value (T). Then all the gray level values below this T will be classified as black (0), and those above T will be white (1). The segmentation problem becomes one of selecting the proper value for the threshold T. A frequent method used to select T is by analyzing the histograms of the type of images that want to be segmented. The ideal case is when the histogram presents only two dominant modes and a clear valley (bimodal). In this case the value of T is selected as the valley point between the two modes. In real applications histograms are more complex, with many peaks and not clear valleys, and it is not always easy to select the value of T.

## 3 AUTOMATIC THRESHOLDS

Automatically selected threshold value for each image by the system without human intervention is called an automatic threshold scheme. This is requirement the knowledge about the intensity characteristics of the objects, sizes of the objects, fractions of the image occupied by the objects and the number of different types of objects appearing in the image

## 4 THRESHOLD TECHNIQUES

Threshold technique is one of the important techniques in image segmentation. This technique can be expressed as:

T=T[x, y, p(x, y), f(x, y]        (1)

Where: T is the threshold value.
x, y are the coordinates of the threshold value point.
p(x,y) ,f(x,y) are points the gray level image pixels.
Threshold image g(x,y) can be define:

$$g(x,y) = \begin{cases} 1 & if \quad f(x,y) > T \\ 0 & if \quad f(x,y) \leq T \end{cases} \quad (2)$$

This paper applied five threshold techniques:

### 4.1 Mean Technique
This technique used the mean value of the pixels as the threshold value and works well in strict cases of the images that have approximately half to the pixels belonging to the objects and the other half to background. This technique case rarely happens.

### 4.2 P-Tile Technique
The p-tile technique uses knowledge about the area size of the desired object to the threshold an image.

The P-tile method is one of the earliest threshold methods based on the gray level histogram [5]. It assumes the objects in an image are brighter than the background, and occupy a fixed percentage of the picture area. This fixed percentage of picture area is also known as P%. The threshold is defined as the gray level that mostly corresponds to mapping at least P% of the gray level into the object.

Let n be the maximum gray level value, H (i) be the histogram of image (i = 0. n), and P be the object area ratio. The algorithm of the P-tile method is as follows:

S=sum (H(i))        (3)
Let f=s
For k=1 to n
f=f-H(k-1)
If (f/t) <p then stop
T=k

Where: S total area of image
f is the initialize all area as object area
T is the final threshold value

This method is simple and suitable for all sizes of objects. It yields good anti-noise capabilities; however, it is obviously not applicable if the object area ratio is unknown or varies from picture to picture [6].

### 4.3 Histogram Dependent Technique (HDT)
The histogram based techniques is dependent on the success of the estimating the threshold value that separates the two homogonous region of the object and background of an image. This required that, the image formation be of two homogonous and will-separated regions and there exists a threshold value that separated these regions.

The (HDT) is suitable for image with large homogonous and will separate regions where all area of the objects and background are homogonous and except the area between the objects and background.

This technique can be expressed as:

C (T) =$P_1$ (T)$\sigma_1^2$(T) +$P_2$(T)$\sigma_2^2$(T)        (4)

Where:
C (T) is the within-group variance.
$P_1$ (T) is the probability for group with values less than T.
$P_2$(T) is the probability for group with values greater than T.
$\sigma_1$ (T) is the variance of group of pixels with values less than or equal T.$\sigma_2$ (T) is the variance of group of pixels with values greater than T.

### 4.4 EMT Technique
The threshold image by using edge maximization technique (EMT) is used when there are more than one homogenous region in image or where there is a change on illumination between the object and its background. In this case portion of the object may be merged with the background or portions of the background may as an object. To this reason any of the automatic threshold selection techniques performance becomes much better in images with large homogenous and well separated regions. This techniques segmentation depend on the research about the maximum edge threshold in the image to start segmentation that image with help the edge detection techniques operators ( for example kiresh operator technique).

### 4.5 Visual Technique
These techniques improve people's ability to accurately search for target items. These techniques are similar to one another P-Tile technique in that they all use the component segments of original images in novel ways to improve visual search performance but it is different from  p-tile don't active when the noise is present in the image .

## 5 EXPERIMENTS VERIFICATIONS

### 5.1 Testing Procedure
The threshold segmentation was implemented using (MATLAB R2007a, 7.4a) and tested the segment techniques on the three images illustrated in the Figure 1.

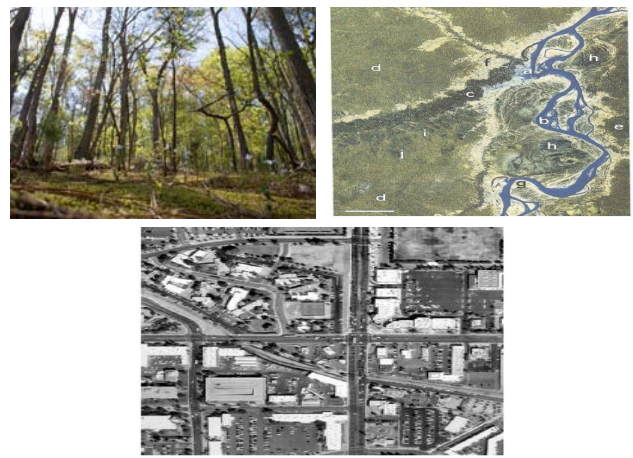

Fig.1. Original Images

Five techniques applied of the threshold secementation: Mean technique, P-Tile technique, Visual technique, HDT technique and EMT technique.

### 5.2 SIMULATION RESULTS

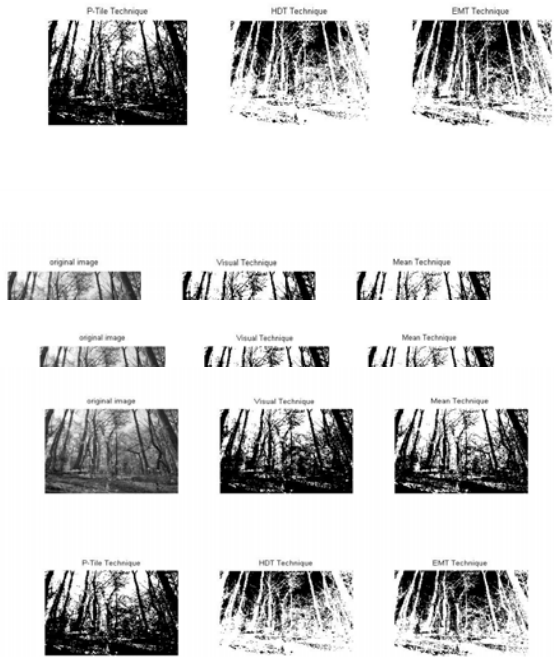

Fig.2. Threshold techniques from left to right original image, Visual technique T=127, Mean Technique, P-Tile technique T=127, HDT Technique and EMT Technique.

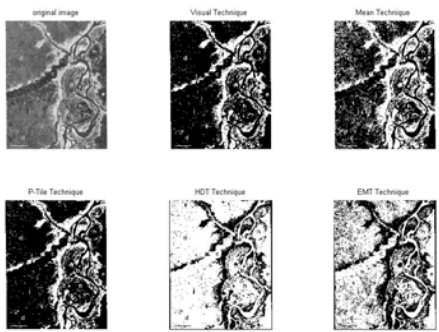

Fig.3. Threshold techniques from left to right original image, Visual technique T=167, Mean Technique, P-Tile technique T=167, HDT Technique and EMT Technique.

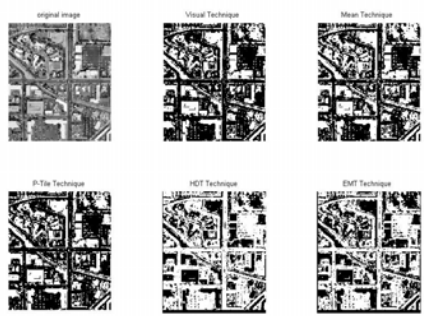

Fig.4. Threshold techniques from left to right original image, Visoual technique T=43, Mean Technique, P-Tile technique T=43, HDT Technique and EMT Technique.

## 6 CONCLUSIONS

In this paper, the comparative studies applied by using five techniques of threshold segmentation techniques image*: Mean method, P-Tile method, Histogram Dependent Technique (HDT), Edge Maximization Technique (EMT) and visual Technique* on the three satellite image illustrated in Fig.1.Acomparative study are explained & experiments are carried out for different techniques HDT and EMT techniques respectively are the best techniques for threshold techniques images. This result can be   seen in the Figures.1, 2and 3.

**REFERENCES**

[1]  Gonzalez and Woods, "Digital image processing", 2nd Edition, prentice hall, 2002.

[2]  Kenneth R. Castelman, "Digital image processing", Tsinghua Univ Press, 2003.

[3]   Shuqian He,  Jiangqun Ni , Lihua Wu , Hongjian Wei , Sixuan Zhao, "Image threshold segmentation method with 2-D histogram based on multi-resolution analysis", Computer Science & Education, ICCSE, 25-28 July 2009, PP.753 – 757, Nanning, China.

[4]  Guang Yang, Kexiong Chen, Maiyu Zhou, Zhonglin Xu, Yongtian Chen, "Study on Statistics Iterative Thresholding Segmentation Based on Aviation Image," snpd, vol. 2, pp.187-188, Eighth ACIS International Conference on Software Engineering, Artificial Intelligence, Networking, and Parallel/Distributed Computing (SNPD 2007), 2007.

[5]   Du  Gen-yuan,Miao  Fang,Tian  Sheng-li,Guo  Xi-rong.,"Remote Sensing Image Sequence Segmentation Based on the Modified Fuzzy C-means", Journal of Software, Vol. 5, No. 1, PP.28-35, 2009.

[6]   A.S. Abutaleb, "Automatic Thresholding of Gray-Level Pictures Using Two Dimensional Entropy", Computer Vision, Graphics, and Image Processing, Vol.47, PP.22-32, 1989.

[7]   J. Kittler and J. Illingworth, "Minimum Error Thresholding", Pattern Recognition, Vol.19, No.1, PP.41-47, 1986.

[8]   K.H. Liang and J.J.W Mao, "Image Thresholding by Minimizing the  Measures  of  Fuzziness", Pattern  Recognition,  Vol.28,  No.1, PP.41-51, 1995.

[9] F. Samopa, A. Asano.,"Hybrid Image Thresholding Method using Edge Detection", IJCSNS International Journal of Computer Science and Network Security, Vol.9 No.4, PP.292-299, April 2009.

**AUTHORS PROFILE**

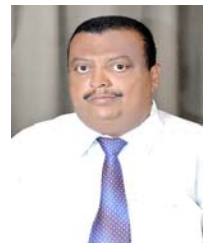

**Mr. Salem Saleh Al-amri.**Received the B.E degree in Mechanical Engineering from University  of  Aden,  Yemen,  1991,  the M.Sc.degree  Computer  science  (IT)  from North Mahrashtra University (N.M.U), India, Jalgaon ,2006, Research student Ph.D in the department of computer science (S.R.T.M.U), India, Nanded.He have seven interntoinal papers, 1 international conference, 2 national conferences.He is lecturer in Minerals & Oil Faculty, aden university, Yemen,He is membership in International Association of Engineers (IAENG).

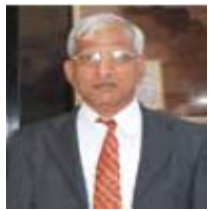

**Dr. N.V. Kalyankar**.B Sc.Maths, Physics, Chemistry, Marathwada University,Aurangabad,India,1978,M.Sc. Nuclear Physics,Marathwada University,Aurangabad,India,1980.Diploma in Higher Education,Shivaji University, Kolhapur, India,1984.Ph.D. in Physics, Dr.B.A.M.University**,** Aurangabad, India,1995.Principal Yeshwant Mahavidyalaya College, Membership of Academic Bodies, Chairman, Information Technology Society State Level Organization, Life Member of Indian Laser Association**,** Member Indian Institute of Public Administration, New Delhi**,** Member Chinmay Education Society, Nanded.He has one publication book, 23 journals papers, two seminars Papers and three conferences papers.

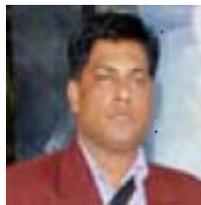

**Dr. S. D Khamitkar.** M. Sc. Ph.D.Computer Science Reader & Director(School of Computational Science)Swami Ramanand Teerth Marathwada University, Nanded**,**14 Years PG Teaching**,** Publications 08 International,Research Guide (10 Students registered),Member Board of Studies(Computer Application),Member Research and RecognitionCommittee (RRC) (Computer Studies).